\documentclass[letterpaper, 10 pt, conference]{ieeeconf}  

\IEEEoverridecommandlockouts                              

\usepackage{amsmath} 
\usepackage{amssymb} 
\usepackage{amsfonts}
\usepackage{mathtools}
\usepackage{times}  
\usepackage{balance}
\usepackage{pdfpages}
\usepackage{fancyhdr}

\usepackage{tabularx}

\usepackage{graphicx} 
\usepackage{blindtext}
\usepackage{hyperref}
\usepackage{multirow}
\usepackage{tabularx}
\usepackage{booktabs}
\usepackage{subfig}
\usepackage{float}
\usepackage{color}
\usepackage{cite}
\usepackage[flushleft]{threeparttable}

\usepackage{xhfill}

\definecolor{axesRed}{RGB}{214,39,40}
\definecolor{axesGreen}{RGB}{44,160,44}
\definecolor{axesBlue}{RGB}{31,119,180}
\definecolor{myGold}{RGB}{255,192,0}

\title{
\LARGE \bf Geometric Structure Aided Visual Inertial Localization
}
\author{Huaiyang Huang, Haoyang Ye, Jianhao Jiao, Yuxiang Sun and Ming Liu
\thanks{$^{1}$Huaiyang Huang, Haoyang Ye, Jianhao Jiao, Yuxiang Sun and Ming Liu are with the Robotics Institute at the Hong Kong University of Science and Technology, Hong Kong.
\texttt{\{hhuangat, hyeab, jjiao, eeyxsun, eelium\}@ust.hk}.}
}

\begin{document}

\newcommand\Tstrut{\rule{0pt}{2.6ex}}         
\newcommand\Bstrut{\rule[-0.9ex]{0pt}{0pt}}   
\newcommand\todo[1]{\textcolor{yellow}{TODO: #1.}}         

\IEEEaftertitletext{\vspace{-1.0\baselineskip}}
\maketitle

\begin{abstract}
    Visual Localization is an essential component in autonomous navigation.
    Existing approaches are either based on the visual structure from SLAM/SfM or the geometric structure from dense mapping.
    To take the advantages of both, in this work, we present a complete visual inertial localization system based on a hybrid map representation
    to reduce the computational cost and increase the positioning accuracy.
    Specially, we propose two modules for data association and batch optimization, respectively.
    To this end,
    we develop an efficient data association module to associate map components with local features, which takes only $2$ms to generate temporal landmarks.
    For batch optimization, instead of using visual factors, we develop a module to estimate a pose prior from the instant localization results to constrain poses.
    The experimental results on the EuRoC MAV dataset demonstrate a competitive performance compared to the state of the arts.
    Specially, our system achieves an average position error in 1.7 cm with 100\% recall.
    The timings show that the proposed modules reduce the computational cost by 20-30\%.
    We will make our implementation open source at \url{http://github.com/hyhuang1995/gmmloc/}.
\end{abstract}

\IEEEpeerreviewmaketitle

\section{Introduction and Related Works}
\label{sec:intro}

Visual Localization is a crucial building block for autonomous navigation, as it provides essential information for other functionalities \cite{liu2013visual}.
In recent years, significant progress has been made for metric visual localization that aims to recover 6-DoF camera poses.
Based on the map representation,
the prevalent pipelines can be categorized into \textit{visual structure}- and \textit{geometric structure}-based methods.

Among them, localizing based on the \textit{visual structure} is the most prevailing paradigm\cite{sattler2016efficient,svarm2016city, sarlin2019coarse, sattler2018benchmarking}.
Generally, the visual structure is represented by a bipartite graph between landmarks and keyframes,
which can be built from Structure-from-Motion (SfM) \cite{schoenberger2016sfm} or Simultaneous Localization and Mapping (SLAM) \cite{mur2017orb}.
Global and local features along with geometric information (e.g., camera poses, landmark positions) are also preserved.
During the localization phase, it retrieves the candidate visual frames in the map and search for correspondences via feature matching.
Camera pose can then be recovered in a Perspective-n-Point (PnP) scheme.
Localization based on the visual structure can resolve a kidnapped localization problem, and the map storage is typically lightweight.
However, it suffers from the camera viewpoint and visual appearance variances \cite{sattler2018benchmarking}.

Recently, \textit{geometric structure}-based methods raise increasing concerns \cite{caselitz2016mloc,kim2018stereo,ding2018laser,huang2019metric,zuo2019visual,ye2020monocular, huang2020gmmloc}.
This track of methods quantize the prior map with dense structure components (e.g., pointcloud, voxels, surfels).
Generally, they do not focus on a kidnapped problem, and instead regard the visual localization problem as a registration problem between local visual structure and global dense map.
The intuition behind is that, compared to visual information, structural information can be more invariant against view-point or illumination change.
However, in these systems, how to bootstrap the localization without a relatively accurate initial guess has not been well considered \cite{caselitz2016mloc}.
In addition, how to efficiently introduce geometric information from dense structure is also an open problem to explore \cite{huang2020gmmloc}.

In this work, we propose a visual-inertial localization system with a hybrid map representation, which takes the advantage of visual and geometric structure and temporal visual information.
Based on this representation, we can easily resolve a kidnapped localization problem without dealing with how to initialize camera poses cross different modalities.
In addition, with the aid of prior geometric structure, we could efficiently generate temporal landmarks, and this process does not require exhaustive inter-frame matching in traditional vision pipelines \cite{schoenberger2016sfm, mur2017orb}, which would be more time-consuming.
The landmark positions can then be recovered and further optimized with geometric constraints.
Another interesting issue is the trade-off between accuracy and time cost.
Unlike a typical SLAM system, optimizing the states related to historical frames is less meaningful for a localization problem, while with the instant localization we are able to achieve a relatively accurate state estimation result.
In this way, how to minimize the computational overhead in batch optimization is also an interesting question to think over.
Therefore, in this work, we propose a windowed optimization scheme with linearized term as prior, which improves the efficiency with a comparative localization accuracy compared to other methods.
The experimental results on a public dataset show that our method can provide an average accuracy around 1.7 cm in a challenging indoor environment.
We summarize our contributions as follows:

\begin{itemize}
    \item We present a visual-inertial localization system based on a hybrid map representation, which takes the advantage from both visual structure and geometric structure.
    \item We propose a map-feature association module for temporal landmark generation, the computational cost of which is reduced and bounded.
    \item We propose an efficient batch optimization scheme with camera poses constrained by the prior terms from the instant localization.
    \item Experimental results show that our system achieves a competitive localization performance while balances the computational cost.
\end{itemize}

\section{System Overview and Preliminaries}

\begin{figure}[t!]
	\centering
	\includegraphics[width=0.48\textwidth]{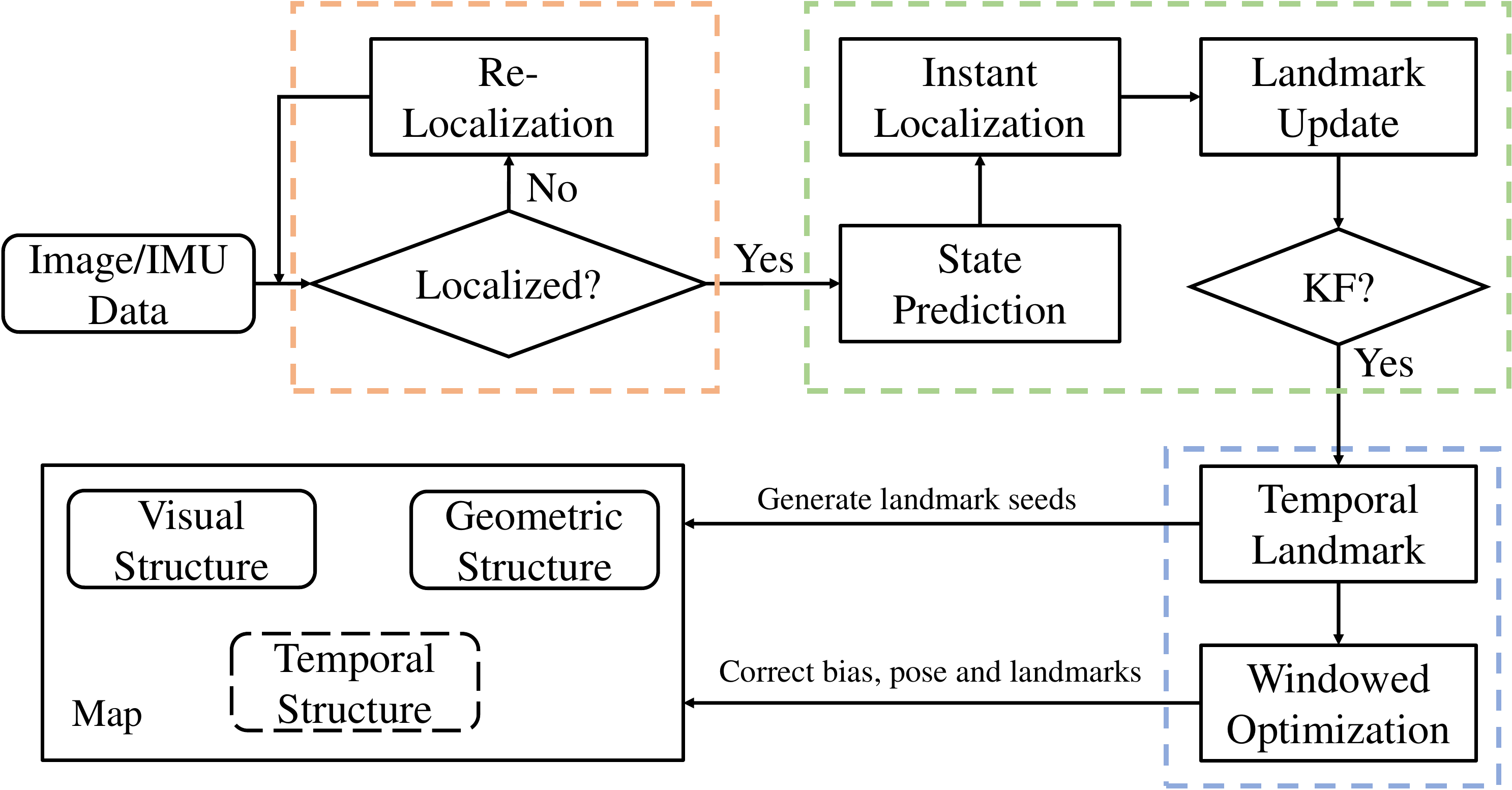}
	\caption{ System overview. KF is the abbreviation for keyframe.}
	\vspace{-1.5em}
	\label{fig.system}
\end{figure}

The proposed visual-inertial localization framework is presented in \autoref{fig.system}.
In the following sections, we briefly introduce different modules along with some preliminaries.

\subsection{Map representation}
We propose to localize cameras with a hybrid map representation, which consists of both \textit{visual structure} and
\textit{geometric structure}.
In our system,
We leverage visual SLAM/SfM methods for the \textit{visual structure} construction,
where the landmarks are filtered by the number of frames that can observe them.
Then, the \textit{geometric structure} is modelled by a Gaussian Mixture Model (GMM), which provides a prior distribution of landmarks:
$p({^W\mathbf{p}_i} | \mathcal{M}) = \prod_j \pi_j p({^W\mathbf{p}_i} | \mathcal{G}_j) = \prod_j \pi_j \mathcal{N}(\boldsymbol{\mu}_j, \boldsymbol{\Sigma}_j)$, with $\pi_j, \boldsymbol{\mu}_j, \boldsymbol{\Sigma}_j$ the weight, mean vector and covariance matrix for the $j$-th component $\mathcal{G}_j$.
In addition, we further voxelize the geometric structure to support the proposed association module between map components and local features, with details in \autoref{sec.local_map}.
With the aid of geometric structure, our system efficiently generates the temporal visual structure to enhance the robustness.

\subsection{Re-localization}
As aforementioned, previous works based on prior geometric information generally require a global initial guess to bootstrap the localization system,
which is difficult to retrieve in a kidnaped problem.
Here we generally follow the pipeline of HLoc \cite{sarlin2019coarse} to re-localize the camera during initialization or tracking lost.
Briefly, the system first retrieves the top-10 keyframe candidates, with NetVLAD \cite{arandjelovic2016netvlad} as the global descriptor.
Then we cluster the candidates by their covisibility, and match local features against each cluster.
With the 2D-3D correspondences, the camera poses is recovered via PnP with RANSAC, and then refined with inlier matches.
However, for the robustness, accuracy and smoothness of the estimated trajectory, we consider that it is a better practice to leverage temporal visual information in localization.
In other words, we explicitly track the matched local features for instant localization rather than performing re-localization at each single frame.

\subsection{State Prediction}
Considering a visual localization problem, the states to estimate each single frame is $\boldsymbol{\xi}_k = \left[\boldsymbol{\phi}_k, \mathbf{t}_k\right]$, where $\boldsymbol{\phi}_k, \mathbf{t}_k$ represent the rotational and translational states, respectively. Then, the rotation matrix from the map frame to the body frame is given as $\mathbf{R}_{B_kW} =\exp(\boldsymbol{\phi}_k^\land)$, and $\mathbf{R}_{B_kW}\in SO(3)$, where the wedge operator $(\cdot)^\land$ turns $\boldsymbol{\phi}_k \in \mathfrak{so}(3)$ into a skew-symmetric matrix.
And we have the translation from the body frame to the world frame $^{B_k}\mathbf{t}_{B_kW} = \mathbf{t}_k$, expressed under the $k$-th body frame.
Using inertial measurements introduces additional states $\left[\mathbf{v}_k, \mathbf{b}^a_k, \mathbf{b}^g_k\right]$, which are the velocity of body frame expressed under the world frame $\mathbf{v}_k = {^W\mathbf{v}_{B_k}}$, the bias term of accelerometer and gyroscope, respectively.
Then the total state vector is denoted by $\mathbf{x}_k = \left[\boldsymbol{\xi}_k, \mathbf{v}_k, \mathbf{b}^a_k, \mathbf{b}^g_k\right]$.
Predicting current states is a common procedure in the state estimation.
In our system, if IMU data is available, the states of the next frame can be predicted via the IMU measurement model.
Otherwise, the constant velocity model is applied for a vision-only implementation.
The relevant descriptions can be found in \cite{hyhuang2020geosup}.

\subsection{Instant Localization}
\label{sec.iloc}
As we stated before, we explicitly utilize sequential information to localize each single frame.
The instant localization module deals with the 2D-3D association, state estimation and covariance recovery.

\subsubsection{Tracking}
Following the tracking scheme in ORB-SLAM2 \cite{mur2017orb}, here we use predicted camera poses as prior information to search for the 2D-3D association to estimate relevant states.
In our implementation, as the global prior information is more reliable, we first try to associate the observed landmarks from the nearest database keyframe.
We also keep this in mind when matching against the previous frame and the (temporal and prior) visual structure, where we give priority to the landmarks in the prior visual structure.

\subsubsection{Pose Estimation}
For the pose recovery, we follow the optimization scheme in \cite{mur2017visual}.
Here we briefly describe it as preliminaries.
With 3D-2D association and optional inertial measurements, current states can then be solved in a least-squares problem,
which seeks to minimize the following objective function:
\begin{equation*}
	E = E_{\text{inertial}}(\mathbf{x}_k, \mathbf{x}_{k-1}) + E_{\text{prior}}(\mathbf{x}_{k-1}) + \sum_{i \in \mathcal{V}_k} E_{\text{proj}}(\boldsymbol{\xi}_{k},
	{^W\mathbf{p}_i)},
\end{equation*}
where $\mathcal{V}_k$ is the set of landmarks visible by the $k$-th frame.
The visual factor is defined by the reprojection error $E_{\text{proj}}(\boldsymbol{\xi}, {^W\mathbf{p}_i}) = \rho(\mathbf{e}^T_\text{proj}(\cdot) \boldsymbol{\Sigma}_{pi} \mathbf{e}_\text{proj}(\cdot))$, with
\begin{equation*}
	\mathbf{e}_\text{proj}(\cdot) \doteq \mathbf{e}_\text{proj}(\boldsymbol{\xi}_k, {^W\mathbf{p}_i}) = \pi ({^{C_k}\mathbf{p}_i}) - \bar{\mathbf{u}}_{ik},
\end{equation*}
\begin{equation*}
	{^{C_k}\mathbf{p}_i} = \mathbf{R}_{CB} (\mathbf{R}_{B_kW}\cdot{^W\mathbf{p}_i} + {^{B_k}\mathbf{t}_{B_kW}}) + {^C\mathbf{t}_{CB}},
\end{equation*}
where $\bar{\mathbf{u}}_{ik}$ represents the local measurement, $\pi$ denotes the camera projection function, $\boldsymbol{\Sigma}_{pi}$ is the covariance matrix for $\bar{\mathbf{u}}_{ik}$, $\mathbf{R}_{CB}, ^{C}\mathbf{t}_{CB}$ are the extrinsic parameters between camera and IMU.
$E_{\text{inertial}}$ and $E_{\text{prior}}$ are the inertial and prior term, respectively.
We use similar formulations as those in \cite{forster2016manifold, mur2017visual} with some trivial modifications. The relevant formulations are described in \cite{hyhuang2020geosup}.
Same as \cite{mur2017visual}, after the optimization, the relevant factors are marginalized out for a prior term of $\mathbf{x}_k$.

\subsubsection{Covariance Recovery}
After the optimization, we are able to approximate the localization covariance from the optimization.
Denote the state we care about as $\mathbf{x}$ and its update as $\mathbf{x} \boxplus \delta \mathbf{x}$.
After an iterative least-squares estimation, we have a normal equation $\mathbf{H}_\mathbf{x} \delta \hat{\mathbf{x}} = \mathbf{b}_{\mathbf{x}}$.
$\mathbf{H}_\mathbf{x}, \mathbf{b}_\mathbf{x}$ might be calculated from the Schur complement, depending on whether $\mathbf{x}$ is the full state vector.
This provides a posterior distribution of the state update $\delta \hat{\mathbf{x}}$:
$\delta\mathbf{x} \sim \mathcal{N}(\delta\hat{\mathbf{x}}, \boldsymbol{\Sigma}_{\delta \mathbf{x}})$, with $\boldsymbol{\Sigma}_{\delta \mathbf{x}} = \mathbf{H}_\mathbf{x}^{-1}$.
Then, the covariance of the state $\mathbf{x}$ can be propagated via:
\begin{equation}
	\boldsymbol{\Sigma}_{\mathbf{x}} = \mathbf{J}(\mathbf{x}, \delta \mathbf{x}) \boldsymbol{\Sigma}_{\delta\mathbf{x}} \mathbf{J}(\mathbf{x}, \delta \mathbf{x})^T \quad \text{with}
\end{equation}
\begin{equation}
	\mathbf{J}(\mathbf{x} , \delta \mathbf{x}) =
	\left.\frac{\partial (\mathbf{x} \boxplus {\delta\mathbf{x}})}{\partial \delta \mathbf{x}} \right|_{\delta \mathbf{x}=\mathbf{0}}
\end{equation}
For $\mathbf{t}, \mathbf{v}, \mathbf{b}_a, \mathbf{b}_g$, the update is simply addition in $\mathbb{R}^3$, so the jacobian is an identity matrix and the covariance remains the same.
On the contrary, for the rotation parameters $\boldsymbol{\phi}$, the update is preformed on tangent space of $SO(3)$, yielding
\begin{equation}
	\mathbf{J}(\boldsymbol{\phi}, \delta \boldsymbol{\phi}) =
	\left. \frac{\partial (\boldsymbol{\phi} \boxplus {\delta\boldsymbol{\phi}})}{\partial \delta \boldsymbol{\phi}}\right|_{\delta \boldsymbol{\phi}=\mathbf{0}}
	= \mathbf{J}_{r}^{-1}(\boldsymbol{\phi})
\end{equation}
with $\mathbf{J}_{r}^{-1}(\boldsymbol{\phi})$ the inverse right jacobian of $\boldsymbol{\phi}$.

The covariance of states, especially rotation and translation, determines how certain the localization result is.
Some detailed explanation and experimental results can be found in \cite{hyhuang2020geosup}.
In addition, this formulation inspires us to replace the visual factors with a pose prior term in the batch optimization, as described in \autoref{sec.opt}.

\subsection{Landmark Creation and Optimization}

If only the prior visual structure is used, the localization performance will be highly dependent on the view overlap and appearance similarity between database images and queries.
In this way, the total quantity of correspondences can be insufficient for recovering accurate states.
Therefore, we propose to generate temporal landmarks with the aid of geometric structure here.
For the state optimization, we leverage a fixed-lag smoother to correct current states and update the positions of temporal landmarks.
To efficiently resolve this batched estimation problem, we propose to use pose prior factors instead of visual factors.
These two modules are described in the \autoref{sec.local_map} and \autoref{sec.opt} with details.

\begin{figure}[t!]
	\centering
	\includegraphics[width=0.48\textwidth]{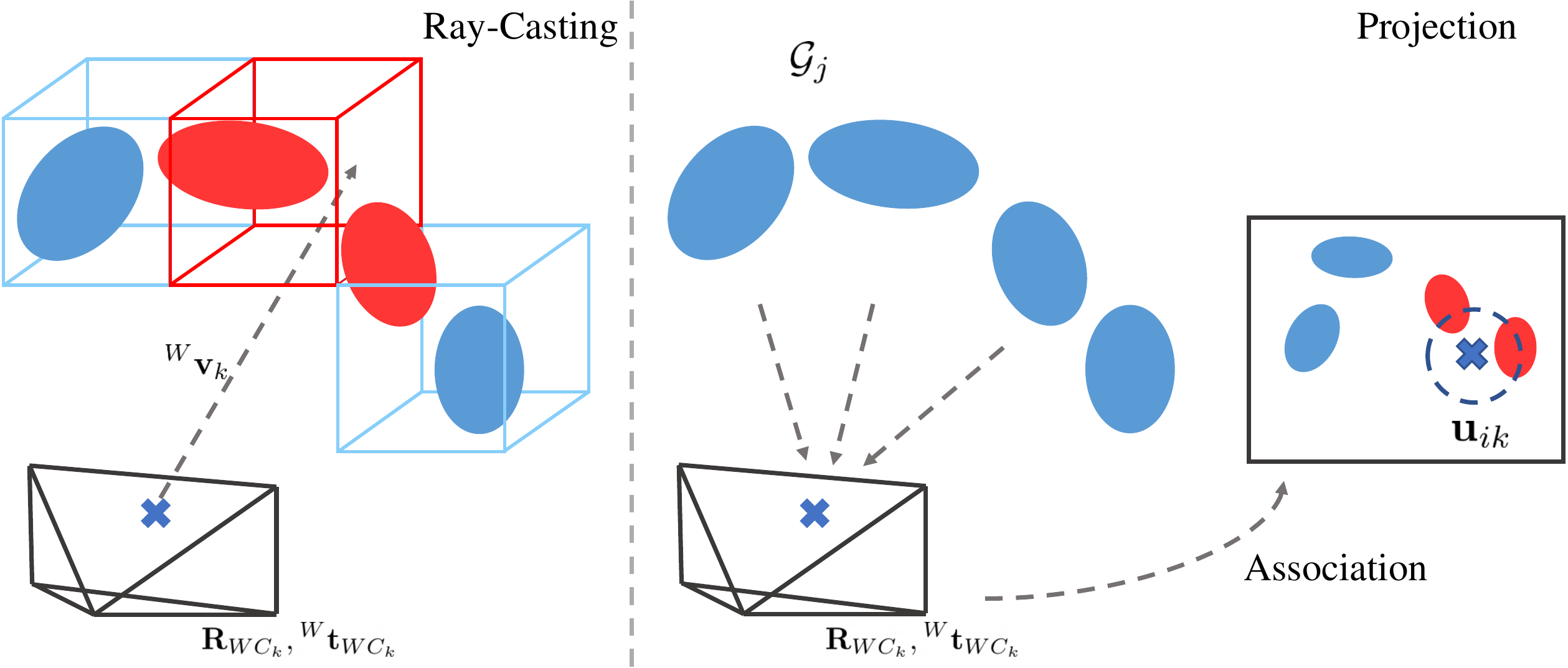}
	\caption{
		Comparative illustration of the \textit{ray-casting} (left) and \textit{projection} (right) method for associating map components and local features. The dashed arrow represents the information flow for generating the association.
	}
	\vspace{-1.5em}
	\label{fig.association}
\end{figure}

\section{Temporal landmark generation}
\label{sec.local_map}

A solution of associating local features with map components is proposed in GMMLoc \cite{huang2020gmmloc}, named as \textit{projection} method here.
We briefly review this association scheme.
GMMLoc first projects all the map components to the image coordinate as 2D distributions by applying non-linear transforms on 3D gaussian distributions.
The local features are then associated with map components in the 2D space.
To filter out occluded components, it sorts the projected components by depth, and compare the depth of each component and its neighborhoods.
The drawbacks of the \textit{projection} method can be three-fold: 1)
when applied to generic camera models, the projection function needs to be linearized, yielding a bad approximation when the camera distortion is large; 2) the computational time is highly dependent on the number of visible components; 3) sorting operation is sequential thus is difficult to be parallelized.

To resolve these issues, in this work, we propose an improved solution that is more invariant to camera model and presents reduced and bounded computational cost.
This method is inspired by techniques widely used in volumetric dense mapping, such as ray-casting, thus we name it \textit{ray-casting} method here.
An comparative illustration of these two methods is shown in \autoref{fig.association}.

As a prerequisite, we extend the GMM into a voxelized representation in the mapping stage.
To this end, we first adopt the \textit{voxel-hashing} method in Voxblox \cite{oleynikova2017voxblox} to divide the Euclidean space into a set of voxels $\left\{V_t\right\}$, with a given resolution $\delta_v = 0.1m$.
Each voxel $V_t$ represents a region $R_{V_t} = \left\{\mathbf{x} | \mathbf{x} \in \mathbb{R}^3, l < \mathbf{c}^T\mathbf{x}< u\right\}$, defined by a box function with parameters $l, u, \mathbf{c}$.
Then for each Gaussian component, we locate their intersected voxels via the cumulative likelihood:
\begin{equation}
	p(V_t | \mathcal{G}_j) = \iiint_{\mathcal{R}_{V_t}}  \exp(-\frac{1}{2}\|\mathbf{p} - \boldsymbol{\mu}_j\|_{\boldsymbol{\Sigma}}) \, d\mathbf{p}.
\end{equation}
We adopt a numerical method \cite{genz2004numerical} to compute this integral as there is no analytical solution.
If $p(V_t | \mathcal{G}_j) > \delta_l = 0.1$, the component $\mathcal{G}_j$ is considered intersected with the voxel $V_t$, and therefore in our system a component can be associated with multiple intersected voxels.

With this voxelized map representation, we then associate the unmatched local features with components in the map.
Given a localized frame, with camera pose $[\mathbf{R}_{WC_k}, {^W\mathbf{t}_{WC_k}}]$ and the local features $\left\{\mathbf{u}_{ki}\right\}_{i=1\dots N}$,
for each candidate feature $\mathbf{u}_{ki}$, we can define a ray parameterized by $\mathbf{r}_{ki} = \left[^W\mathbf{v}_{ki}, {^W\mathbf{t}_{WC_k}}\right]$, with the bearing vector under world coordinate $^W\mathbf{v}_{ki}$, which is given by:
${^{W}\mathbf{v}_{ki}} = \mathbf{R}_{WC_k} \pi^{-1}(\mathbf{u}_{ki})$,
and ${^W\mathbf{t}_{WC_k}}$ the origin of the ray.
Then we can parameterize the position of landmark observed by $\mathbf{u}_{ki}$ as $^W\mathbf{p}_i = {^W\mathbf{t}_{WC_k}} + \lambda {^W\mathbf{v}_{ki}}$, where $\lambda$ is the depth along the ray.
To find out proper map association for $\mathbf{u}_{ki}$, we cast the ray into the voxelized map.
For each voxel along the ray, we examine the components intersected with it and the metric is defined by Mahalanobious distance between a line and a component ${dist}\left(\mathbf{r}_{ki}, \mathcal{G}_j\right)$, which is explained as follows:
decomposing the covariance of $\mathcal{G}_j$ using SVD, we have:
\begin{equation*}
	\boldsymbol{\Sigma}_j = \mathbf{R}_j\mathbf{S}_j\mathbf{R}_j^T,
	\mathbf{S}_j = \text{diag}(\lambda_1, \lambda_2, \lambda_3), \mathbf{R}_j = [\mathbf{e}_{j1}, \mathbf{e}_{j2}, \mathbf{e}_{j3}].
\end{equation*}
Applying the whitening transform, the ray parameters become \cite{lu2008robust}:
\begin{equation}
	\label{eq.whiten}
	\begin{aligned}
		{\mathbf{t}^{\prime}_{WC_i}} & = \mathbf{S}_j^{-\frac{1}{2}} \mathbf{R}_j^T ( {^{W}\mathbf{t}_{WC_i}} - {^{W}\boldsymbol{\mu}_{j}}), \\
		{\mathbf{v}^{\prime}_{ki}}   & = \mathbf{S}_j^{-\frac{1}{2}} \mathbf{R}_j^T \cdot {^{W}\mathbf{v}_{kj}},
	\end{aligned}
\end{equation}
with which the distance metric is given by:
\begin{equation}
	\begin{aligned}
		{dist}\left(\mathbf{r}_{ki}, \mathcal{G}_j\right) & = \frac{1}{\|\mathbf{v}_{ki}^{\prime}\|} \|\mathbf{t}_{WC_i}^{\prime} \times \mathbf{v}_{ki}^{\prime}\|. \\
	\end{aligned}
\end{equation}
For enhancing the efficiency, $\mathbf{S}_j^{-\frac{1}{2}}\mathbf{R}_j^T$ in \autoref{eq.whiten} can be pre-computed for each component $\mathcal{G}_j$ by the Cholesky decomposition: $\boldsymbol{\Sigma}_j^{-1} = \mathbf{LL}^T = (\mathbf{S}_j^{-1/2}\mathbf{R}_j^T)^T(\mathbf{S}_j^{-1/2}\mathbf{R}_j^T)$.

With this metric, for each component associated with the same voxel, we adopt $\mathcal{X}^2$-test for the component that has the closest distance to the optical ray, which gives a threshold $\delta_r = 2.8$.
And if the distance is within $\delta_r$, we associate the component with the feature $\mathbf{u}_{ki}$.
Finally, with the association, the depth $\lambda$ in the landmark position parameterization can be trivially recovered by:
$ \lambda = {\mathbf{e}_1^T (\boldsymbol{\mu} - \mathbf{t}_{WC})}/{\mathbf{e}_1^T\mathbf{v}}$.
We only perform this operation for local features that are not associated with landmarks in the map.
We also sort them by detection score and only keep the top-100 keypoints to bound the computational cost.
These strategies are kept the same for comparing the \textit{projection} method and the \textit{ray-casting} method in \autoref{sec:experiments}.
Unlike the \textit{projection} method, time complexity of the \textit{ray-casting} method is mainly dependent on number of local features for querying,
hence is more bounded.
`'
We name these initialized points as \textit{seed}s, which are not immediately used for localization.
Instead we further verified them in the \textit{Landmark Update} step, with subsequent visual information along with localized camera poses.
Similar to the matching scheme in \autoref{sec.iloc}, we match the correspondences with the localized camera pose as prior.
Briefly, if a \textit{seed} is observed with sufficient parallax over multiple frames, it will be activated as a landmark for the future localization.
In the implementation, the minimum parallax is set to $4$-px and the minimum number of observation is set to $4$.
\textit{Seed}s not visible in the current window will be cleared after each update.

\begin{figure}[t!]
	\centering
	\includegraphics[width=0.48\textwidth]{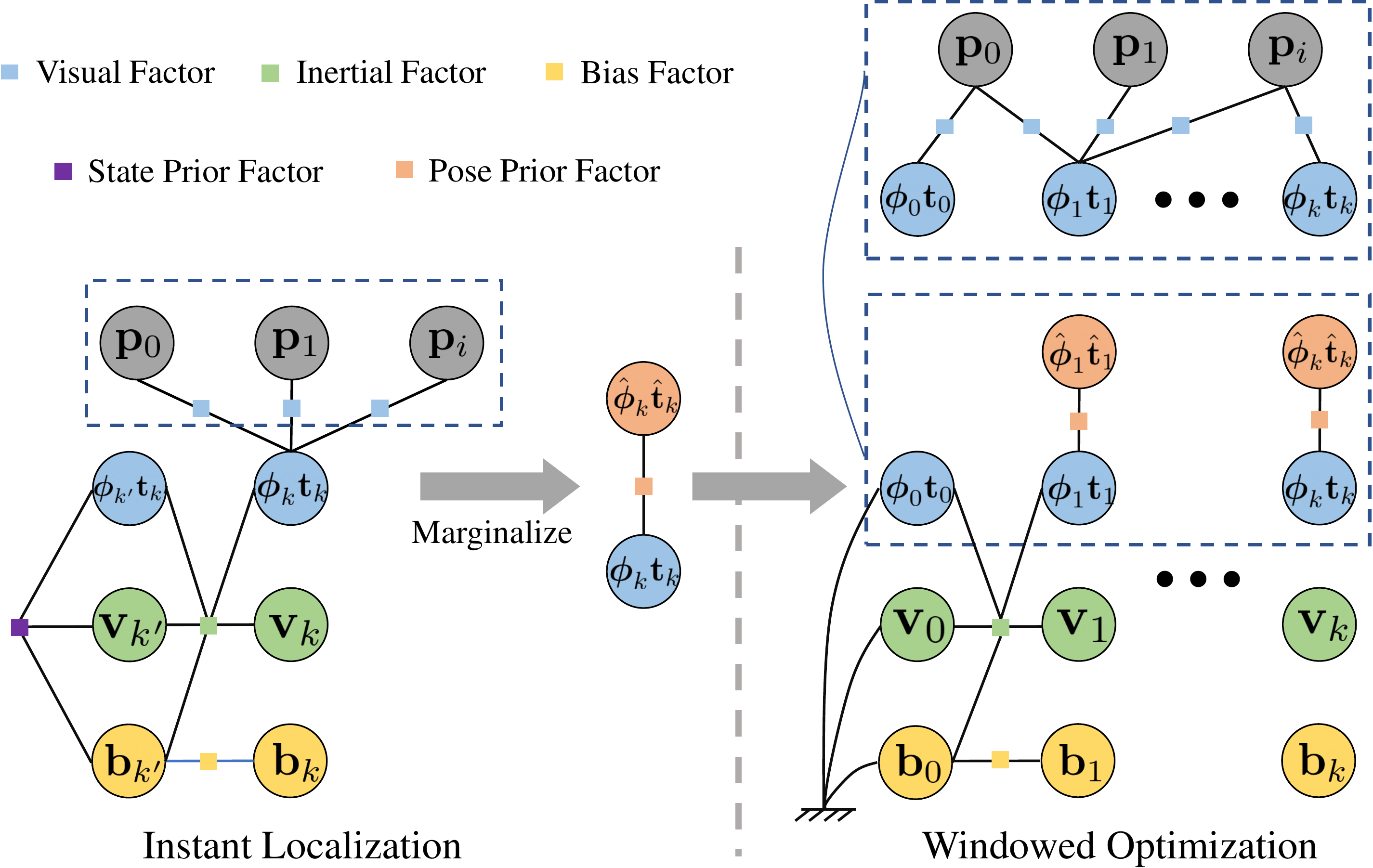}
	\caption{Factor graph representation for the windowed optimization.
		After instant localization, we marginalize out the visual factors for a prior distribution of camera pose.
		In the motion-only BA, instead of using non-linear visual factors, we use this pose prior as constraints to correct current states.
	}
	\vspace{-1.5em}
	\label{fig.win_opt}
\end{figure}

\section{Optimization with Global Pose Prior}
\label{sec.opt}

Although we can already recover the global camera pose with the instant localization, some of the states are not able to be properly estimated, especially when the inertial measurements are incorporated.
For example, we find that in practice, $\mathbf{b}^a$ is difficult to converge without batch optimization, yielding a biased state prediction.
Here we proposed a batched yet efficient optimization structure, which can be solved within $10$ms in average.

Firstly, inspired by SVO \cite{forster2014svo}, we decoupled the full Bundle Adjustment (BA) into motion- and structure-only BA,
and \autoref{fig.win_opt} shows the factor graph representation of our motion-only BA pipeline.
For the motion-only BA, we can optimize the camera pose via minimizing
\begin{equation}
	\label{eq.motion_ba}
	E = \sum_{k\in \mathcal{F}} E_{\text{inertial}}(\mathbf{x}_k, \mathbf{x}_{k-1}) + \sum_{k\in \mathcal{F}} \sum_{i \in \mathcal{V}_k} E_\text{proj}(\boldsymbol{\xi}_k, {^W\mathbf{p}_i}),
\end{equation}
where $\mathcal{F}$ is the keyframes in the current window, and we set $\left|\mathcal{F}\right|=15$.
Although the decoupled scheme accelerate the optimization,
we found that iteratively re-linearizing the visual factors can be less effective in computation, while the the instant localization can already provide a relatively accurate estimation.
As a consequence, we propose to marginalize the visual factors as a pose prior instead of using the original visual factors.
After the pose optimization described in \autoref{sec.iloc}, we have the normal equation: $\mathbf{H} \delta \hat{\mathbf{x}}_k = \mathbf{b}$. By reordering the states, this normal equation can be re-written as:
\begin{equation}
	\left[
		\begin{matrix}
			\mathbf{H}_{tt} & \mathbf{H}_{tr} \\
			\mathbf{H}_{rt} & \mathbf{H}_{rr} \\
		\end{matrix}
		\right]
	\left[
		\begin{matrix}
			\delta \hat{\boldsymbol{\xi}}_k \\
			\delta \hat{\mathbf{x}}_r       \\
		\end{matrix}
		\right]
	=
	\left[
		\begin{matrix}
			\mathbf{b}_t \\
			\mathbf{b}_r \\
		\end{matrix}
		\right],
\end{equation}
where $t$ denotes the block of related to the camera pose update $\delta \boldsymbol{\xi}_k$ and $r$ the block of other variables.
The hessian block $\mathbf{H}_{xx}$ can be decomposed by $\mathbf{H}_{xx} = \mathbf{H}_{i} + \mathbf{H}_{p}$, where $\mathbf{H}_i$ and $\mathbf{H}_{p}$ are the hessians related to inertial and visual factors, respectively.
We have $\mathbf{H}_p = \sum_{i \in \mathcal{V}_k} \mathbf{H}_{pi}$ and $\mathbf{H}_{pi}$ is given by:
\begin{equation*}
	E_\text{proj} (\boldsymbol{\xi}_k \boxplus {\delta \boldsymbol{\xi}_k}, {^W\mathbf{p}_i}) \approx \mathbf{J}_{pi}^T \boldsymbol{\Sigma}_{pi}^{-1} \delta \boldsymbol{\xi} + \delta \boldsymbol{\xi}^T \underbrace{ \mathbf{J}_{pi}^T \boldsymbol{\Sigma}_{pi}^{-1}\mathbf{J}_{pi}}_{\mathbf{H}_{pi}} \delta \boldsymbol{\xi},
\end{equation*}
where $\mathbf{J}_{pi}$ is the jacobian of $\mathbf{e}_\text{proj}(\boldsymbol{\xi}_k, {^W \mathbf{p}_i})$ with respect to $\boldsymbol{\xi}_k$.
Marginalizing out the visual factors simply forms a distribution on the camera pose update $\delta \boldsymbol{\xi}_k \sim
	\mathcal{N}(\delta \hat{\boldsymbol{\xi}}_k, \boldsymbol{\Sigma}_{\delta \boldsymbol{\xi}_k}), \boldsymbol{\Sigma}_{\delta \boldsymbol{\xi}_k} = \mathbf{H}_p^{-1}$.

\begin{figure*}[t!]
	\centering
	\includegraphics[width=0.24\textwidth]{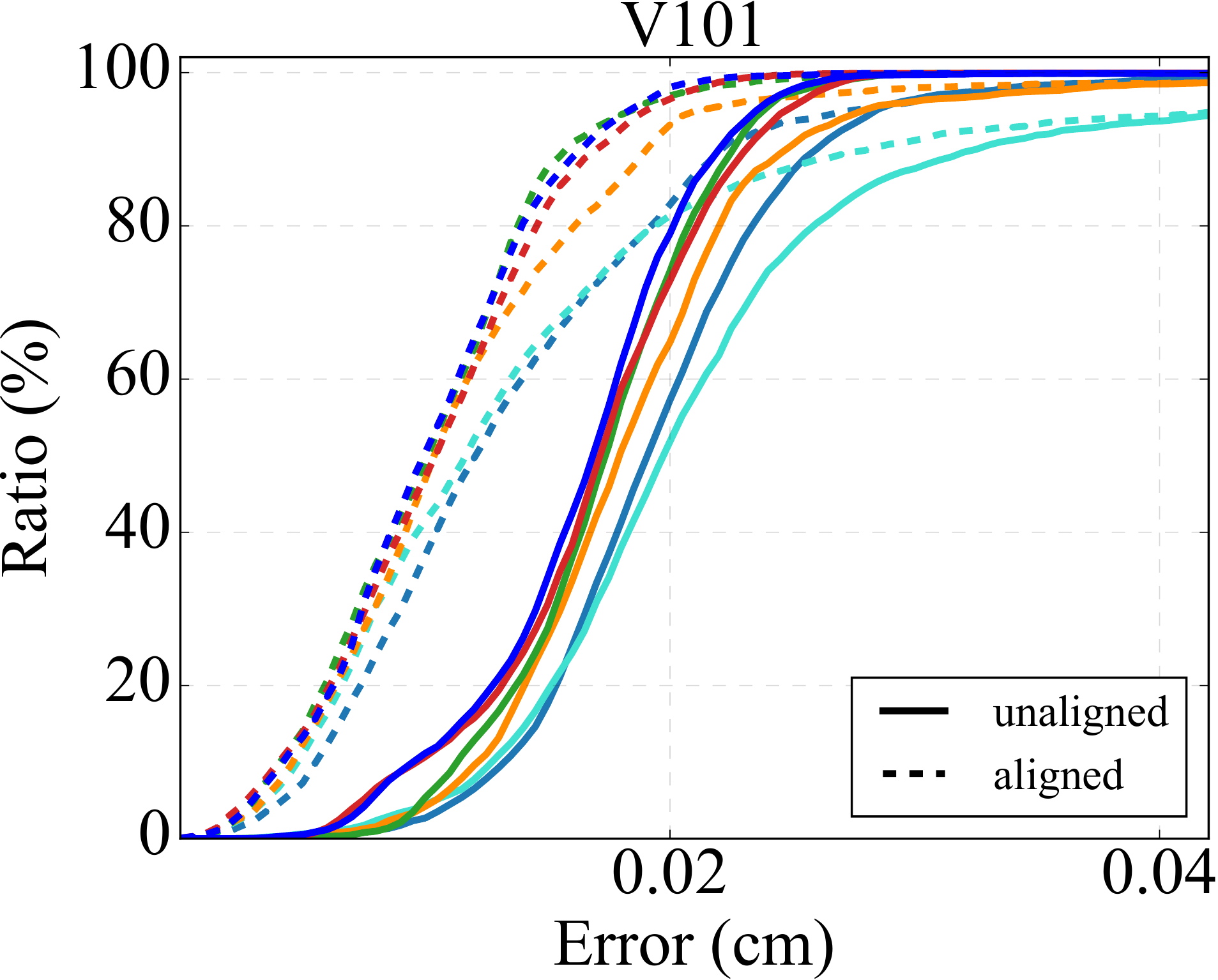}
	\includegraphics[width=0.24\textwidth]{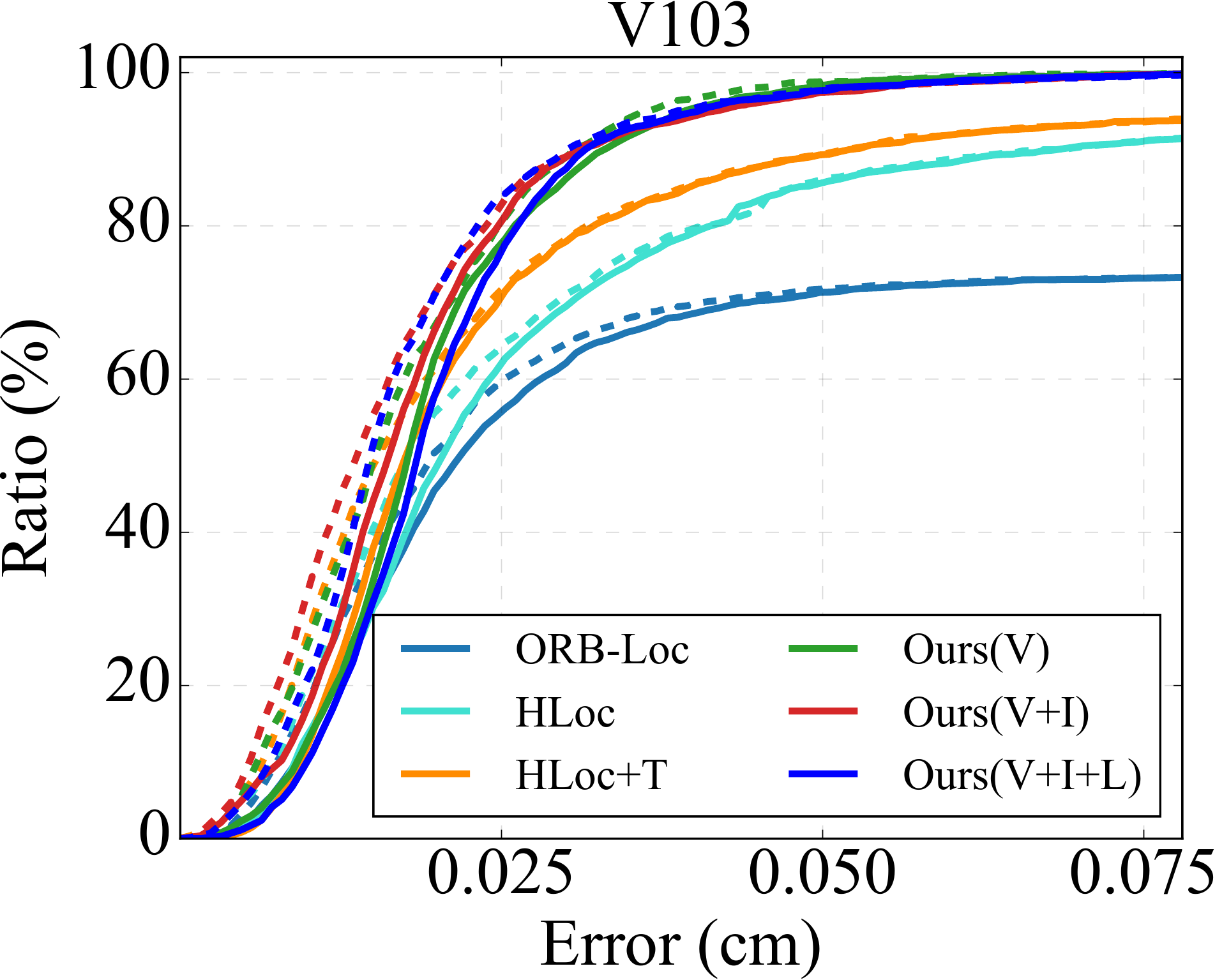}
	\includegraphics[width=0.24\textwidth]{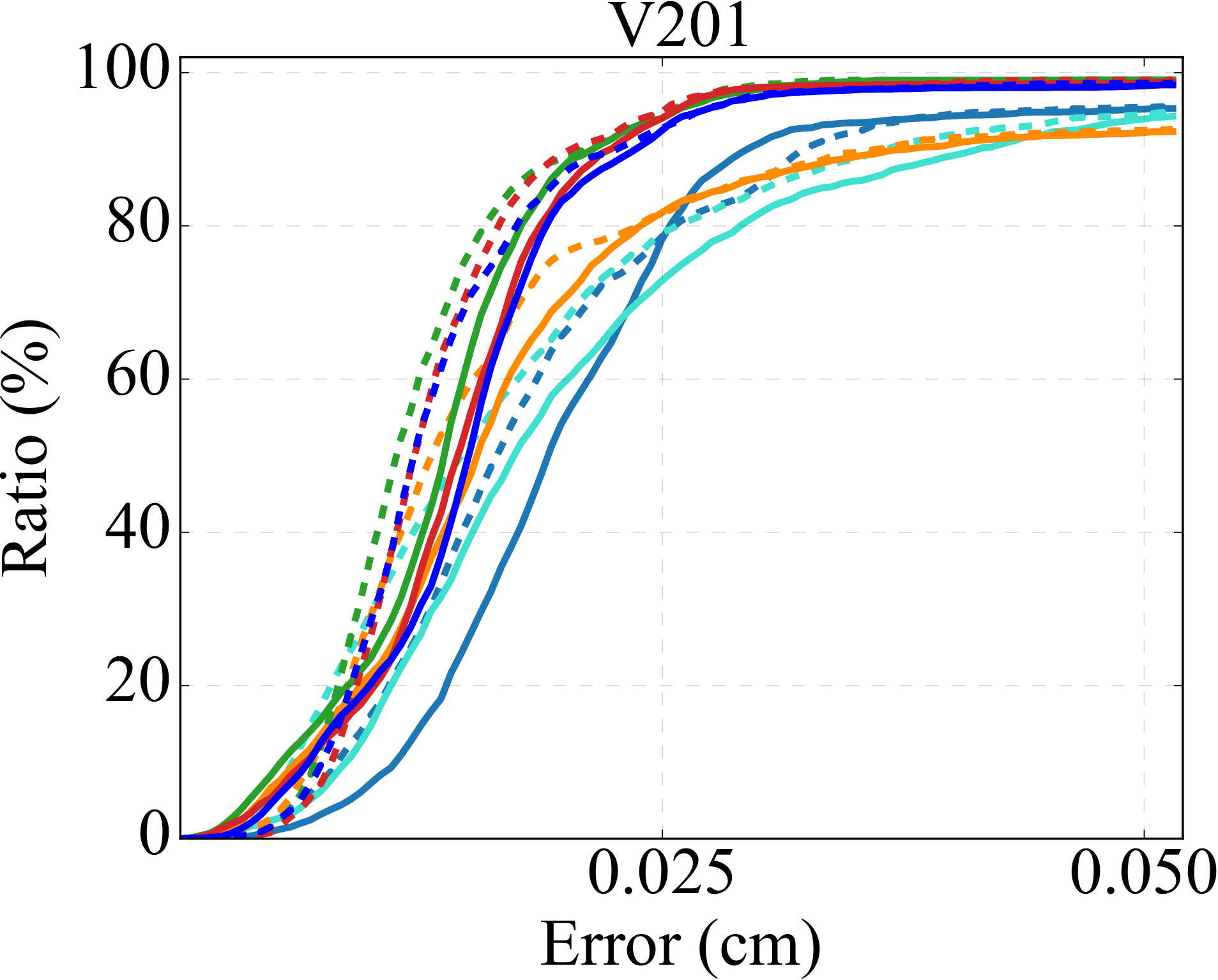}
	\includegraphics[width=0.24\textwidth]{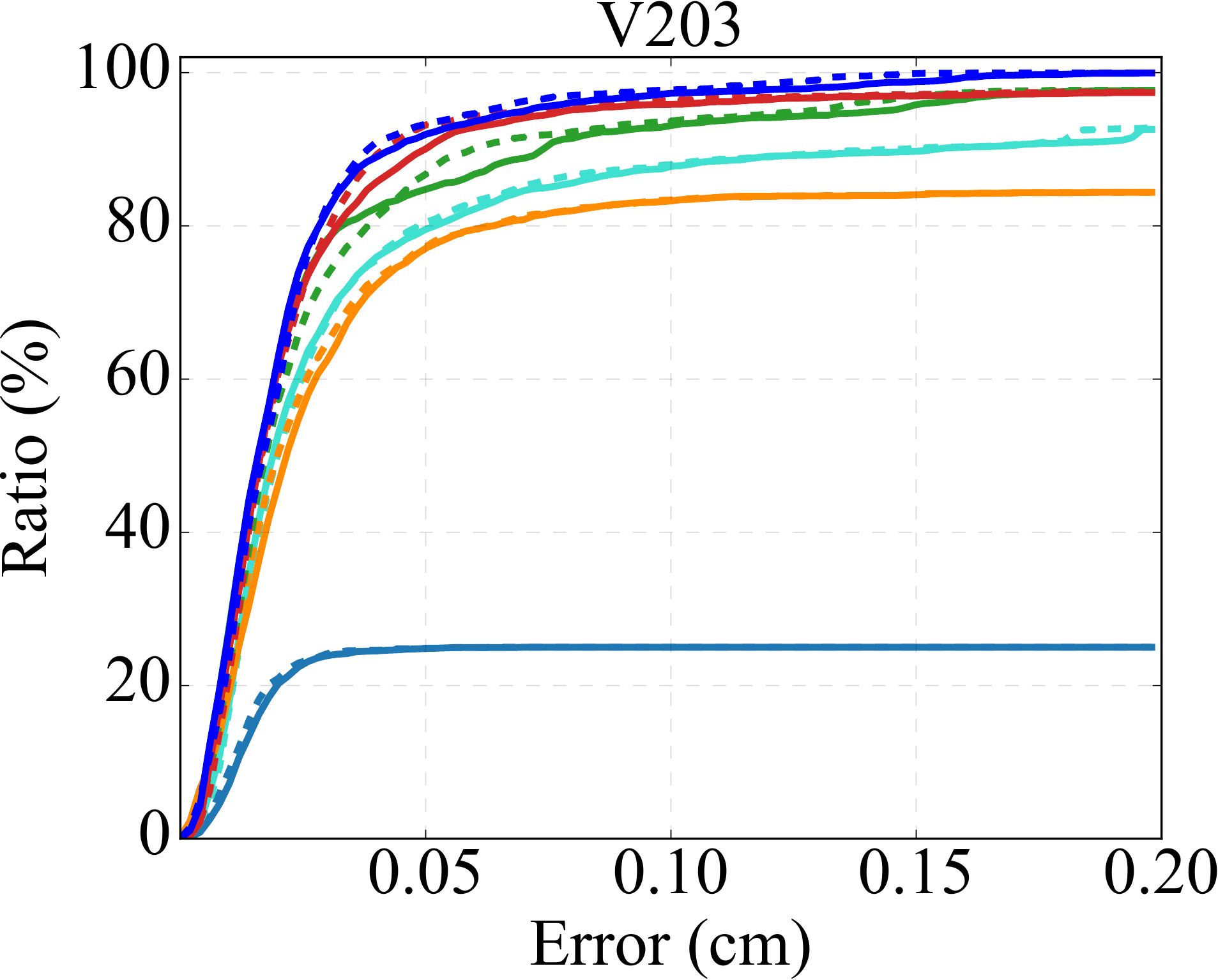}

	\caption{\textbf{Cumulative ratio of APEs} on EuRoC MAV dataset.
		We provide evaluation results with and without alignment against the groundtruth, as described in the legend.
	}
	\vspace{-1.5em}
	\label{fig.ratio_loc_err}
\end{figure*}

With this prior information, we can formulate the prior pose constraints as:
\begin{equation}
	\begin{aligned}
		\mathbf{e}_{\text{rot}}   & = \log (\hat{\mathbf{R}}_{WB_k}\mathbf{R}_{B_kW}\exp(\delta \hat{\boldsymbol{\phi}}^\land))^\lor, \\
		\mathbf{e}_{\text{trans}} & = ({^{B_k}\mathbf{t}_{B_kW}} - {^{B_k}\hat{\mathbf{t}}_{B_kW}}) - \delta\hat{\mathbf{t}}.
	\end{aligned}
\end{equation}
Assuming the pose estimation reaches the local optimal, we have the update of state $\delta \hat{\boldsymbol{\xi}}_k \rightarrow \mathbf{0}$, thus the prior error state can be formulated as
\begin{equation*}
	\mathbf{e}_{\text{pose}}(\boldsymbol{\xi}_k) = [\log (\hat{\mathbf{R}}_{WB_k}\mathbf{R}_{B_kW})^\lor, 
	{^{B_k}\mathbf{t}_{B_kW}} - {^{B_k}\hat{\mathbf{t}}_{B_kW}}].
\end{equation*}
Then the total objective function \autoref{eq.motion_ba} is changed into:
\begin{equation*}
	E = \sum_{k \in \mathcal{F}} E_{\text{inertial}}(\mathbf{x}_k, \mathbf{x}_{k-1}) + \sum_{k \in \mathcal{F}} \rho(\|\mathbf{e}_\text{pose}\left(\boldsymbol{\xi}_k\right)\|_{\boldsymbol{\Sigma}_{\delta \boldsymbol{\xi}_k}}),
\end{equation*}
This problem can be solved by the Levenberg-Marquarelt method.
For the structure-only optimization, similar to GMMLoc \cite{huang2020gmmloc}, we re-triangulate the temporal landmarks with geometric constraints.

The proposed optimization structure is quite similar to Sliding Window Filter (SWF) commonly used in batched state estimation.
A little bit different is that we fix the states relevant to the oldest frame, instead of introducing a prior term from marginalization, same as in \cite{mur2017visual}.
As in the localization system, the global observations are available for most of the time, we believe such information loss is acceptable.
Therefore, we choose this strategy that is much easier for implementation.

\begin{table}[t!]
	\caption{Evaluation on general localization performance on EuRoC MAV dataset.
		We report the mAPE (cm) and the recall (\%) for different methods.
		The three rows represent variants of our method, visual structure-based methods and geometry structure-based methods.
	}
	\label{tab.euroc_acc}
	\centering
	\begin{tabular}{ @{}l c c c c@{} }
		\toprule
		\textbf{Method} & \textbf{V101} & \textbf{V103} & \textbf{V201} & \textbf{V203} \\ 
		\midrule
		Ours (V)        & 1.1 (100.0)   & 1.8 (100.0)   & 1.5 (99.8)    & 2.7 (97.3)    \\
		Ours (V+I)      & 1.1 (100.0)   & 1.7 (100.0)   & 1.5 (100.0)   & 2.6 (99.2)    \\
		Ours (V+I+P)    & 1.2 (100.0)   & 1.9 (100.0)   & 2.8 (100.0)   & 2.4 (100.0)   \\
		Ours (V+I+L)    & 1.1 (100.0)   & 1.8 (100.0)   & 1.6 (100.0)   & 2.3 (100.0)   \\ \midrule
		ORB-Loc         & 1.5 (100.0)   & 1.9 (73.8)    & 1.9 (95.7)    & 1.4 (25.0)    \\
		HLoc            & 1.9 (99.2)    & 2.6 (95.9)    & 2.1 (99.2)    & 3.4 (94.2)    \\
		HLoc + T        & 1.3 (99.2)    & 2.3 (96.9)    & 1.6 (93.6)    & 2.4 (84.6)    \\ \midrule
		DSL             & 3.5 (-)       & 4.5 (-)       & 2.6 (-)       & 10.3 (-)      \\
		GMMLoc          & 3.0 (-)       & 4.7 (-)       & 1.8 (-)       & 5.6 (-)       \\
		MSCKF (w/ map)  & 5.6 (-)       & 8.7 (-)       & 6.9 (-)       & 14.9 (-)      \\
		\bottomrule
	\end{tabular}
	\vspace{-1.5em}
\end{table}

\begin{figure}[t!]
	\centering
	\includegraphics[width=0.48\textwidth]{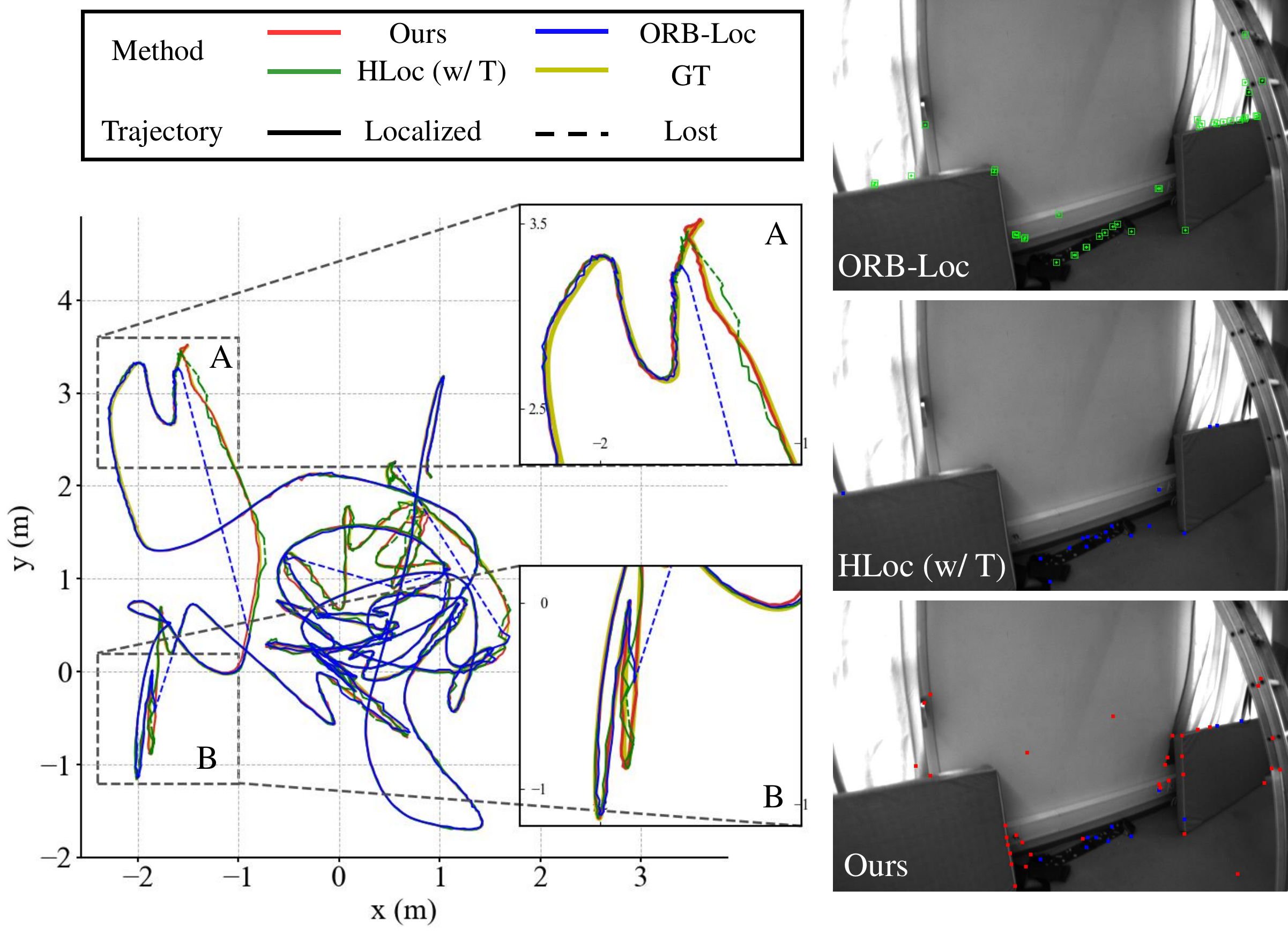}
	\caption{Left: compare reconstructed trajectories against ground truth, with lost part shown as dashed line.
		We show that our method is more robust and is able to estimate a more smooth trajectory locally.
		Right: excerpts in region A for ORB-Loc, HLoc + T and Ours, respectively. With the aid of dense structure, our system generates temporal landmarks (Red) to keep sufficient observations.
	}
	\vspace{-2.0em}
	\label{fig.qualitative}
\end{figure}
\section{Experimental Results}
\label{sec:experiments}
\subsection{Experiment Setup}
In the experiments, we first evaluate different variants of our system, generally named as \textbf{Ours(*)}.
For different variants, we have \textbf{V} standing for the basic vision-only system,
\textbf{I} for the usage of inertial measurements,
\textbf{L} for the usage of pose prior term in the optimization (\autoref{sec.opt}),
and \textbf{P} for replacing the \textit{ray-casting} method with the \textit{projection} method (\autoref{sec.local_map}).
We then evaluate several state-of-the-art methods as baselines, which includes Hierarchical Localization (\textbf{HLoc}) \cite{sarlin2019coarse} and ORB-SLAM2 in localization mode (\textbf{ORB-Loc}) \cite{mur2017orb}.
Considering that HLoc deals with a kidnapped localization problem with single-shot image as input, we extend it with instant localization module same as ours.
This baseline is named as \textbf{HLoc + T}.
In addition, we also report results of several methods based mainly on geometric structure for comparison, i.e., \textbf{DSL} \cite{ye2020monocular}, \textbf{GMMLoc} \cite{huang2020gmmloc} and \textbf{MSCKF (w/ map)} \cite{zuo2019visual}.

For the evaluation, we use EuRoC MAV dataset \cite{burri25012016} that contains two indoor scenes with three sequences on each.
It provides stereo images, IMU data streams along with a dense point cloud map.
The GMM is built with scikit-learn tookit\footnote{\url{https://scikit-learn.org/}}.
In our evaluations, for each scene, we build a visual structure using COLMAP \cite{schoenberger2016sfm} on the medium sequences (V102, V202) and test the performance on the easy (V101, V201) and difficult sequences (V103, V203).
And the visual map is filtered by the minimum number of observations (20 in the experiments).
The same model is used for Ours(*), HLoc, HLoc + T.
We found that the sequence V203 is extremely difficult for the challenging motion profile, illumination condition and missing in the left images, as a consequence of which we use the right image as input and keep the original SfM model for localization.
The experiments have been run on a desktop with an Intel Core i7-8700K CPU, 32GB RAM and NVIDIA 1080Ti GPU.



\subsection{Results and Discussion}

We report the mean Absolute Pose Error (mAPE) and pose recall in \autoref{tab.euroc_acc} and \autoref{fig.ratio_loc_err}.
Generally, the proposed system achieves competitive accuracy and robustness in all the four sequences, as shown in \autoref{tab.euroc_acc}.
While Ours (V+I+L) achieves an average position error of 1.7cm, it also never fails on any of the sequences.
The pose recall curves in \autoref{fig.ratio_loc_err} also show that for most of the cases, our method outperforms the baselines in both fine- and coarse-precision regimes.

Comparing the results from Ours (V) and Ours (V+I), we observe that the performance is very close with some improvements on the difficult sequences (V103 and V203).
As the motion profile on V203 is very aggressive, without proper state prediction from IMU, we observe a recall drop by $1.9\%$.
On the contrary, on easy sequences, we observe slight accuracy drop (e.g., from the distribution curve of V201), as using IMU introduces more parameters to estimate .
These results show that inertial measurements would be more helpful when the situation is more challenging.
Comparing the results from Ours (V+I) and Ours (V+I+P), we observe that with association module being replaced by projection-based method, generally the localization accuracy remains close, while an obvious drop about 1cm occurs on V201.
For Ours (V+I+L), we observe that the overall performance is also close to those without using pose prior.
To our surprise, Ours (V+I+L) performs better than all the other variants on V203.

For methods-based on visual structure, while the results show that ORB-Loc has a similar performance to other methods in two easy sequences, it degrades on the difficult sequences.
This shows that the selection of local and global features can be a significant factor in visual localization.
As HLoc + T is similar to our method in system level, it should be a competitive baseline.
While the localization accuracy is close, the recall drop shows that it is less robust than our method for most of the cases.
In addition, the recall drop compared to HLoc on V201 and V203 is mainly because we do not attempt to re-localize a frame if it is not successfully localized in the instant localization.

For the methods based on geometric structure (DSL, GMMLoc, MSCKF (w/ map)), we observe that they generally has a less accurate localization performance.
This is mainly because the global constraints are formulated similar to point-to-plane distance, which suffer from scene structure degeneration.

\autoref{fig.qualitative} show the reconstructed trajectories of several methods.
Although the localization accuracy of different methods in \autoref{tab.euroc_acc} are close, we observe that generally the trajectory of our methods is better aligned with the ground truth.
As our method benefits from the temporal visual generation, more features can be associated with landmarks, which improves the localization accuracy and robustness.
This also validates the benefit of geometry structure.

\newcommand{\tabline}{\hspace{1em}\xrfill{0.5pt}\hspace{1em}}

\begin{figure}[t!]
	\centering
	\includegraphics[width=0.48\textwidth]{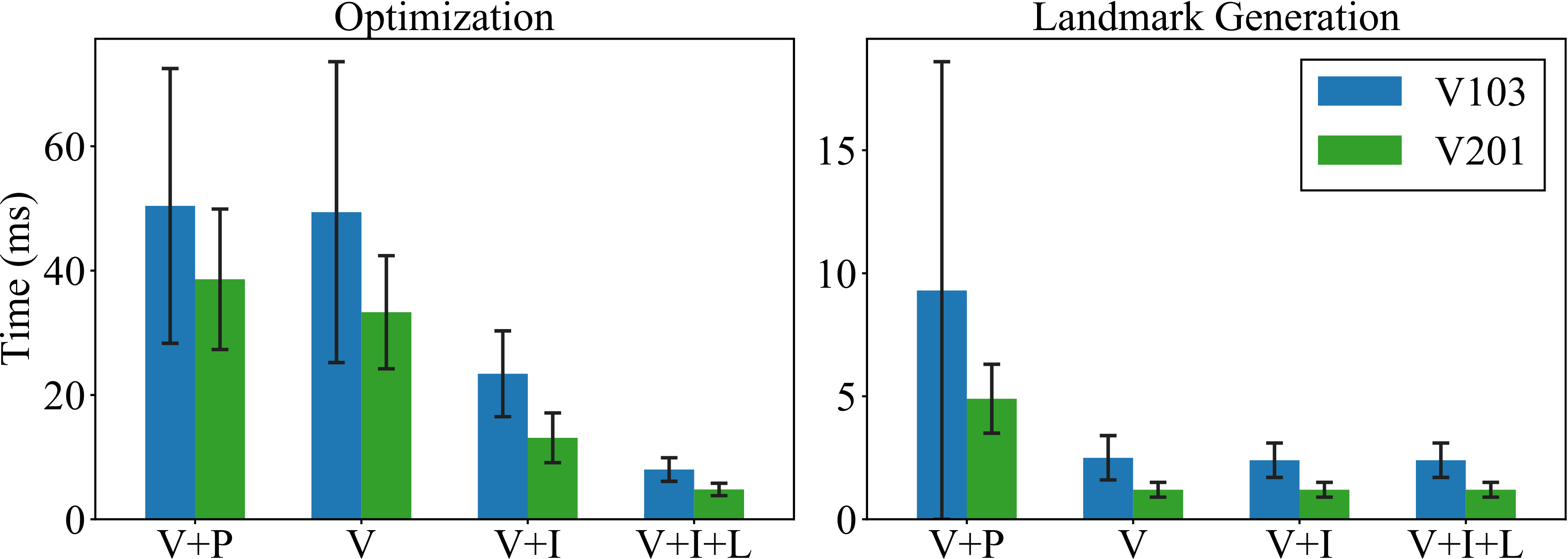}
	\caption{
		Timing results of two modules \textit{optimization} and \textit{landmark generation}, respectively.
	}
	\vspace{-2.0em}
	\label{fig.timing}
\end{figure}

As the proposed localization solution is developed with the consideration of computational constraints, we further evaluate the runtime performance of different variants of our method.
To validate the effectiveness of the proposed modules, we focus on analyse the timings on \textit{temporal landmark creation} and \textit{optimization}, which are reported in \autoref{fig.timing}.
Comparing the two association methods, we observe that that the proposed one significantly improves the efficiency.
In addition, the timing of \textit{projection} method varies a lot (a standard derivation of 9.6ms in V201), which shows that the computational cost of \textit{projection} method is highly dependent on the visible components.
On the contrary, timings for the proposed is more consistent.
Comparing the optimization time, firstly there is a significant improvement when the full BA is decoupled as motion-only and structure-only BA, above ${1}/{2}$ of the time comparing Ours(V) with Ours(V+I).
Furthermore, with camera pose prior from instant localization, there is no need to reconstruct the jacobians related to the visual factors, the optimization time can be further improved by ${2}/{3}$.
More detailed timing results on V201 and V103 are reported in \cite{hyhuang2020geosup}.
In general, timings of Ours(V+I+L) show that we achieve a real-time performance in single-thread (28.8ms on V201 and 38.0ms on V103).




\section{Conclusions}

In this work, we have presented a visual inertial localization system based on a hybrid map representation.
To increase the localization robustness, we have proposed an association method to efficiently create temporal landmarks.
In the windowed optimization, we have proposed a strategy that reduce the computational time of full BA from about $50$ms to under $10$ms.
The experimental results on EuRoC MAV dataset show that how our method can provide a $1-3$ cm localization accuracy with less computational resources in indoor scenarios. In the future, we would like to further improve the robustness and accuracy at the system level.
Although our optimization module is already very efficient,
we believe that solving this problem incrementally would be more suitable in practice, as the measurements typically comes in order.
In addition, as the current feature extraction module is relatively tiaccording to the ame-consuming, distillation from the current model can be another interesting problem to deal with.

\newpage
\balance
\bibliography{main}
\bibliographystyle{IEEEtran}

\newpage


\section*{Supplementary material}

# Geometric Structure Aided Visual Inertial Localization
# —Supplemental Materials—

Huaiyang Huang, Haoyang Ye, Jianhao Jiao, Yuxiang Sun and Ming Liu

## A. State Definition, Update and Prediction

We first recap the state definitions. As in the paper, the state vector for $k$-th frame is defined by:

$$\mathbf{x}_k = [\boldsymbol{\phi}_k, \mathbf{t}_k, \mathbf{v}_k, \mathbf{b}_k^g, \mathbf{b}_k^a] \tag{1}$$

where the rotation and translation of the body frame are:

$$\mathbf{R}_{B_k W} = \exp(\boldsymbol{\phi}_k^\wedge), \quad {}^{B_k}\mathbf{t}_{B_k W} = \mathbf{t}_k \tag{2}$$

In our implementation, the update of states is defined by:

$$\mathbf{x}_k \leftarrow \mathbf{x}_k \boxplus \delta\mathbf{x}_k = \begin{bmatrix} \exp(\boldsymbol{\phi}_k^\wedge)\exp(\delta\boldsymbol{\phi}_k^\wedge) \\ \mathbf{t}_k + \delta\mathbf{t}_k \\ \mathbf{v}_k + \delta\mathbf{v}_k \\ \mathbf{b}_k^g + \delta\mathbf{b}_k^g \\ \mathbf{b}_k^a + \delta\mathbf{b}_k^a \end{bmatrix} \tag{3}$$

We apply the method proposed in [1] to convert raw inertial measurements into pre-integrated factors $\Delta\mathbf{R}_{kk+1}$, $\Delta\mathbf{v}_{kk+1}$ and $\Delta\mathbf{t}_{kk+1}$. With inertial pre-integration, the camera pose of incoming frame $\mathbf{R}_{WB_{k+1}}, {}^W\mathbf{t}_{B_{k+1}}$ can be predicted by:

$$\begin{aligned}
\mathbf{R}_{WB_{k+1}} &= \mathbf{R}_{WB_k}\Delta\mathbf{R}_{kk+1}\exp(\mathbf{J}_{\Delta R}^g \delta\mathbf{b}_k^g), \\
{}^W\mathbf{v}_{B_{k+1}} &= {}^W\mathbf{v}_{B_k} + \mathbf{g}_W\Delta t_k \\
&\quad + \mathbf{R}_{WB_k}\left(\Delta\mathbf{v}_{kk+1} + \mathbf{J}_{\Delta v}^g \delta\mathbf{b}_k^g + \mathbf{J}_{\Delta v}^a \delta\mathbf{b}_k^a\right), \\
{}^W\mathbf{t}_{B_{k+1}} &= {}^W\mathbf{t}_{B_k} + {}^W\mathbf{v}_{B_k}\Delta t_k + \frac{1}{2}{}^W\mathbf{g}\Delta t_k^2 \\
&\quad + \mathbf{R}_{WB_k}\left(\Delta\mathbf{t}_{kk+1} + \mathbf{J}_{\Delta p}^g \delta\mathbf{b}_k^g + \mathbf{J}_{\Delta p}^a \delta\mathbf{b}_k^a\right),
\end{aligned} \tag{4}$$

where $\mathbf{J}_{\Delta R}^g\ \mathbf{J}_{\Delta v}^g\ \mathbf{J}_{\Delta v}^a\ \mathbf{J}_{\Delta t}^g\ \mathbf{J}_{\Delta t}^a$ are the jacobians of pre-integrated terms w.r.t the bias update $\mathbf{b}_k^g$ and $\mathbf{b}_k^a$. For vision-only system, we apply constant-velocity model to predict the current pose:

$$\begin{aligned}
\mathbf{R}_{WB_{k+1}} &= \mathbf{R}_{WB_k}(\mathbf{R}_{B_{k-1}W}\mathbf{R}_{WB_k}) \\
{}^W\mathbf{t}_{WB_{k+1}} &= 2{}^W\mathbf{t}_{WB_k} - {}^W\mathbf{t}_{WB_{k-1}}
\end{aligned} \tag{5}$$

## B. Residual Definition and Jacobians

The residual terms used in the implementation are defined as follows: for the visual factors, we define the residual as the reprojection error:

$$\mathbf{e}_{\text{proj}} = \pi({}^{C_k}\mathbf{p}_i) - \bar{\mathbf{u}}_{ik}, \tag{6}$$

$${}^{C_k}\mathbf{p}_i = \mathbf{R}_{CB}(\mathbf{R}_{B_k W}{}^W\mathbf{p}_i + {}^{B_k}\mathbf{t}_{B_k W}) + {}^C\mathbf{t}_{CB}, \tag{7}$$

where ${}^{C_k}\mathbf{p}_i$ is the landmark position under frame $C_k$. $\bar{\mathbf{u}}_{ik}$ is the local observation for ${}^W\mathbf{p}_i$.

The inertial residual is defined by:

$$\mathbf{e}_{\text{inertial}} = \left[\mathbf{e}_{\text{rot}}^T, \mathbf{e}_{\text{pos}}^T, \mathbf{e}_{\text{vel}}^T, \mathbf{e}_{\text{bias}}^T\right]^T, \tag{8}$$

[1]Huaiyang Huang, Haoyang Ye, Jianhao Jiao, Yuxiang Sun and Ming Liu are with the Robotics Institute at the Hong Kong University of Science and Technology, Hong Kong. *(Corresponding author: Ming Liu.)*
{hhuangat, hyeab, jjiao, eeyxsun, eelium}@ust.hk.

where we have

$$\begin{aligned}
\mathbf{e}_{\text{rot}} &= \log\left(\left(\Delta\mathbf{R}_{kj}\exp(\mathbf{J}_{\Delta R}^g \mathbf{b}_g^j)\right)\mathbf{R}_{B_k W}\mathbf{R}_{WB_j}\right)^\vee, \\
\mathbf{e}_{\text{vel}} &= \mathbf{R}_{B_j W}\Delta\mathbf{v}_{jk} - \left(\Delta\tilde{\mathbf{v}}_{jk} + \mathbf{J}_{\Delta v}^g \delta\mathbf{b}_g^j + \mathbf{J}_{\Delta v}^a \delta\mathbf{b}_a^j\right), \\
\mathbf{e}_{\text{pos}} &= \mathbf{R}_{B_j W}\Delta\mathbf{t}_{jk} - \left(\Delta\tilde{\mathbf{t}}_{jk} + \mathbf{J}_{\Delta t}^g \delta\mathbf{b}_g^j + \mathbf{J}_{\Delta t}^a \delta\mathbf{b}_a^j\right), \\
\mathbf{e}_{\text{bias}} &= \mathbf{b}^j - \mathbf{b}^k,
\end{aligned} \tag{9}$$

with $\Delta\mathbf{v}_{jk}$ and $\Delta\mathbf{t}_{jk}$ given by:

$$\Delta\mathbf{v}_{jk} = {}^W\mathbf{v}_{B_k} - {}^W\mathbf{v}_{B_j} - {}^W\mathbf{g}\Delta t_{jk} \tag{10}$$

$$\Delta\mathbf{t}_{jk} = {}^W\mathbf{t}_{B_k} - {}^W\mathbf{t}_{B_j} - {}^W\mathbf{v}_{B_j}\Delta t_{jk} - \frac{1}{2}{}^W\mathbf{g}\Delta t_{jk}^2 \tag{11}$$

where $\Delta\tilde{\mathbf{R}}_{kj}, \Delta\tilde{\mathbf{v}}_{kj}, \Delta\tilde{\mathbf{t}}_{kj}$ are from the pre-integration, $\mathbf{J}_{\Delta R}^g\ \mathbf{J}_{\Delta v}^g\ \mathbf{J}_{\Delta v}^a\ \mathbf{J}_{\Delta t}^g\ \mathbf{J}_{\Delta t}^a$ are the jacobians related to bias update. This formulation is similar to [1], [2].

The jacobians of different residual terms w.r.t different states are derived as:

$$\frac{\partial \mathbf{e}_{\text{proj}}}{\partial \delta\mathbf{x}_k} = \frac{\partial \mathbf{e}_{\text{proj}}}{\partial {}^{C_k}\mathbf{p}_i} \cdot \frac{\partial {}^{C_k}\mathbf{p}_i}{\partial \delta\mathbf{x}_k}$$

$$\frac{\partial {}^{C_k}\mathbf{p}_i}{\partial \delta\boldsymbol{\phi}_k} = -\mathbf{R}_{CB}\mathbf{R}_{BW} \cdot {}^W\mathbf{p}_i^\wedge, \quad \frac{\partial {}^{C_k}\mathbf{p}_i}{\partial \delta\mathbf{t}_k} = \mathbf{R}_{CB},$$

$$\frac{\partial \mathbf{e}_{\text{rot}}}{\partial \delta\boldsymbol{\phi}_k} = \mathbf{J}_r^{-1}(\mathbf{e}_{\text{rot}})\mathbf{R}_{B_k W}, \quad \frac{\partial \mathbf{e}_{\text{rot}}}{\partial \delta\boldsymbol{\phi}_j} = -\mathbf{J}_r^{-1}(\mathbf{e}_{\text{rot}})\mathbf{R}_{B_j W},$$

$$\frac{\partial \mathbf{e}_{\text{rot}}}{\partial \delta\boldsymbol{\phi}_k} = \mathbf{J}_r^{-1}(\mathbf{e}_{\text{rot}})\mathbf{R}_{B_k W}, \quad \frac{\partial \mathbf{e}_{\text{rot}}}{\partial \delta\boldsymbol{\phi}_j} = -\mathbf{J}_r^{-1}(\mathbf{e}_{\text{rot}})\mathbf{R}_{B_j W},$$

$$\frac{\partial \mathbf{e}_{\text{rot}}}{\partial \delta\mathbf{b}_g^j} = \mathbf{J}_r^{-1} \cdot (\mathbf{e}_{\text{rot}})\mathbf{R}_{B_j W}\mathbf{R}_{WB_k} \cdot \mathbf{J}_r\left(\mathbf{J}_{\Delta R}^g \mathbf{b}_g^j\right),$$

$$\frac{\partial \mathbf{e}_{\text{vel}}}{\partial \delta\boldsymbol{\phi}_j} = -\mathbf{R}_{B_j W} \cdot \Delta\mathbf{v}_{jk}^\wedge, \quad \frac{\partial \mathbf{e}_{\text{vel}}}{\partial \delta\mathbf{v}_j} = \mathbf{R}_{B_j W},$$

$$\frac{\partial \mathbf{e}_{\text{vel}}}{\partial \delta\mathbf{v}_k} = -\mathbf{R}_{B_j W}, \quad \frac{\partial \mathbf{e}_{\text{vel}}}{\partial \delta\mathbf{b}_g^j} = -\mathbf{J}_{\Delta v}^g, \quad \frac{\partial \mathbf{e}_{\text{vel}}}{\partial \delta\mathbf{b}_a^j} = -\mathbf{J}_{\Delta v}^a,$$

$$\frac{\partial \mathbf{e}_{\text{pos}}}{\partial \delta\boldsymbol{\phi}_j} = -\mathbf{R}_{B_j W} \cdot \Delta\mathbf{t}_{jk}^\wedge, \quad \frac{\partial \mathbf{e}_{\text{pos}}}{\partial \delta\boldsymbol{\phi}_k} = \mathbf{R}_{B_j W} \cdot \mathbf{t}_{WB_k}^\wedge,$$

$$\frac{\partial \mathbf{e}_{\text{pos}}}{\partial \delta\mathbf{t}_k} = -\mathbf{R}_{B_j W}\mathbf{R}_{B_k W}^T, \quad \frac{\partial \mathbf{e}_{\text{pos}}}{\partial \delta\mathbf{t}_j} = \mathbf{I}_{3\times 3},$$

$$\frac{\partial \mathbf{e}_{\text{pos}}}{\partial \delta\mathbf{v}_j} = -\mathbf{R}_{B_j W}\Delta t_{jk}, \quad \frac{\partial \mathbf{e}_{\text{pos}}}{\partial \delta\mathbf{b}_g^j} = -\mathbf{J}_{\Delta t}^g, \quad \frac{\partial \mathbf{e}_{\text{pos}}}{\partial \delta\mathbf{b}_a^j} = -\mathbf{J}_{\Delta t}^a.$$

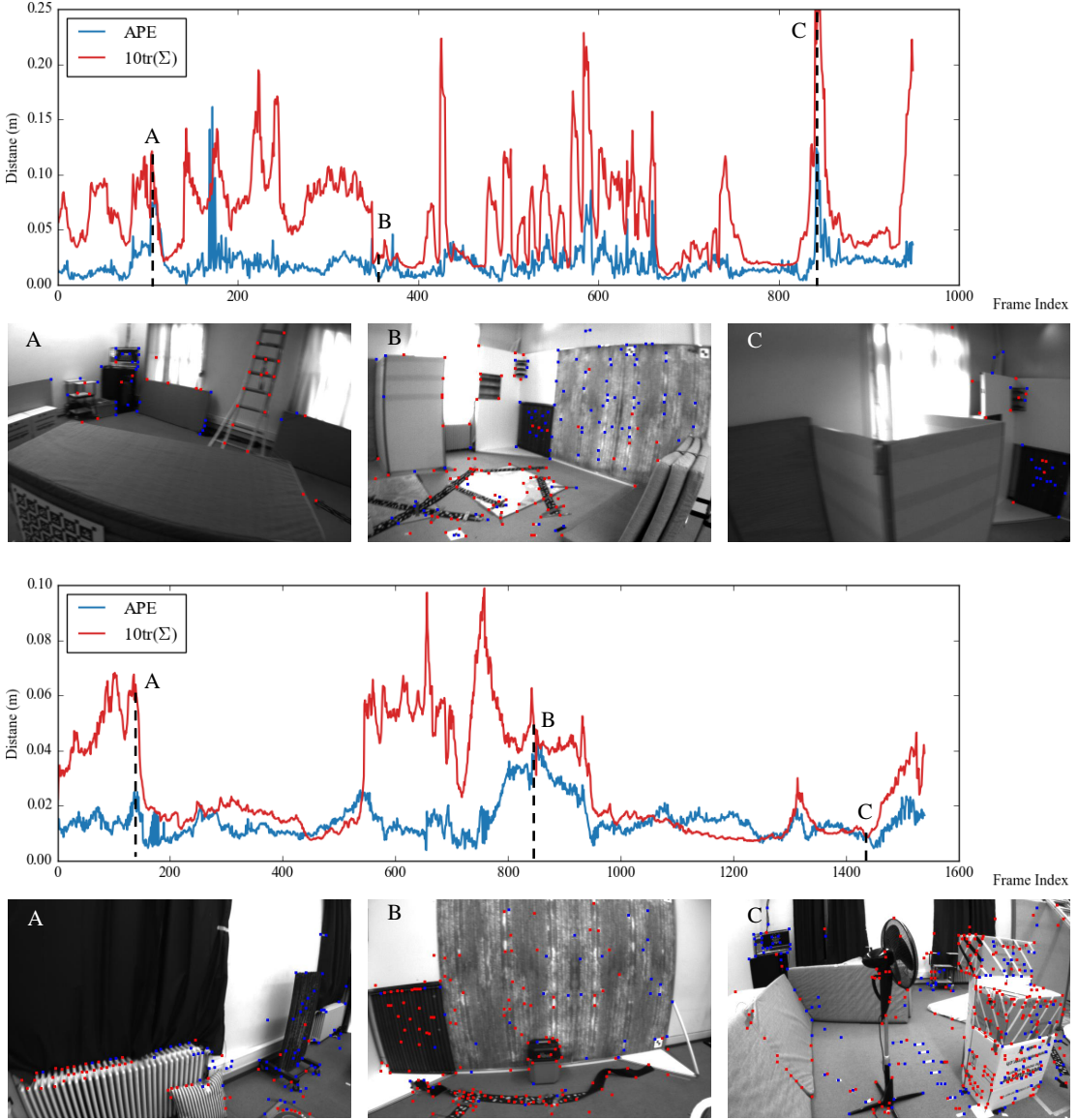

Fig. 1: The trace of covariance and Absolute Pose Error (APE) on V103 (top) and V201 (bottom), along with some excerpts that show local features matched with landmarks in both temporal (shown in red) and prior (shown in blue) structures.

### C. Covariance Recovery

*1) Explanation:* In the manuscript, we provide a general description for covariance recovery, here we give a more detailed explanation along with some example results. Assuming that we would like to recover the covariance of $^W\mathbf{t}_{WB_k}$. In the instant localization, we estimate relevant states via the following optimization problem:

$$\hat{\mathbf{x}}_k, \hat{\mathbf{x}}_{k-1} = \arg\min_{\mathbf{x}_k, \mathbf{x}_{k-1}} E_{\text{inertial}}(\mathbf{x}_k, \mathbf{x}_{k-1}) + E_{\text{prior}}(\mathbf{x}_k) + \sum_{i \in \mathcal{V}_k} E_{\text{proj}}(\boldsymbol{\xi}_k, {}^W\mathbf{p}_i),$$

which resulting in solving the following normal equation:

$$\mathbf{H}\delta\mathbf{x} = \mathbf{b}, \tag{12}$$

where $\mathbf{H}$ and $\mathbf{b}$ are the information matrix and vector, $\delta\mathbf{x}$ is the update of the state vector, respectively. To recover the covariance of $^W\mathbf{t}_{WB_k}$, we first reorder the state vector as $[\mathbf{t}_k, \mathbf{x}_r]$, $\mathbf{x}_r$ is for the rest states, then we have:

$$\begin{bmatrix} \mathbf{H}_{tt} & \mathbf{H}_{tr} \\ \mathbf{H}_{rt} & \mathbf{H}_{rr} \end{bmatrix} \begin{bmatrix} \delta\mathbf{t}_k \\ \delta\mathbf{x}_r \end{bmatrix} = \begin{bmatrix} \mathbf{b}_t \\ \mathbf{b}_r \end{bmatrix}. \tag{13}$$

By applying schur complement, we have

$$\bar{\mathbf{H}}_{tt}\delta\mathbf{t}_k = \bar{\mathbf{b}}_t, \tag{14}$$

with

$$\bar{\mathbf{H}}_{tt} = \mathbf{H}_{tt} - \mathbf{H}_{tr}\mathbf{H}_{rr}^{-1}\mathbf{H}_{rt}, \quad \bar{\mathbf{b}}_t = \mathbf{b}_t - \mathbf{H}_{tr}\mathbf{H}_{rr}^{-1}\mathbf{b}_r. \tag{15}$$

With *Proposition. 1* and *Proposition. 2*, Eq. 14 forms a prior distribution on $\delta\mathbf{t}_k$, $\delta\mathbf{t}_k \sim \mathcal{N}(\delta\hat{\mathbf{t}}_k, \boldsymbol{\Sigma}_{tt})$, with $\boldsymbol{\Sigma}_{tt} = \bar{\mathbf{H}}_{tt}^{-1}$. To obtain the covariance of translation under map frame, i.e., $^W\mathbf{t}_{WB_k}$, we can simply propagate the

covariance via:

$$\Sigma_t = \mathbf{R}_{WB} \Sigma_{tt} \mathbf{R}_{BW} \quad (16)$$

*Proposition 1.* With the measurement noise modelled by Gaussian, in a least-squares problem, the covariance of states can be approximated by $\Sigma = \mathbf{H}^{-1}$ when reaching optimal.

*Proof 1.* Given the measurement function as:

$$\mathbf{z} = h(\mathbf{x}) + \mathbf{n}_z, \quad \mathbf{n}_z \sim \mathcal{N}(\mathbf{0}, \Sigma_z). \quad (17)$$

When the estimation reaching the optimal, we can linearize it as:

$$\mathbf{z} = h(\hat{\mathbf{x}}) + \mathbf{J}\delta\mathbf{x}, \quad \text{with} \quad \mathbf{J} = \frac{\partial h(\hat{\mathbf{x}} \boxplus \delta\mathbf{x})}{\partial \delta\mathbf{x}}. \quad (18)$$

By propagating the uncertainty, we have:

$$\Sigma_z = \mathbf{J}\Sigma\mathbf{J}^T. \quad (19)$$

Apply inversion on both side, we have:

$$\Sigma_z^{-1} = (\mathbf{J}\Sigma\mathbf{J}^T)^{-1}, \quad (20)$$

thus we have $\Sigma = (\mathbf{J}^T \Sigma_z^{-1} \mathbf{J})^{-1} = \mathbf{H}^{-1}$.

*Proposition 2.* By applying Schur Complement to recover state covariance, the above scheme forms a marginal distribution, given as:

$$p(\delta\mathbf{t}_k \mid \mathbf{z}_{rt}) = \int_{\delta\mathbf{x}_r} p(\delta\mathbf{t}_k, \delta\mathbf{x}_r \mid \mathbf{z}_{rt}) d\delta\mathbf{x}_r \approx \mathcal{N}(\delta\hat{\mathbf{t}}_k, \Sigma_{tt})$$

where the joint distribution of $[\delta\mathbf{t}_k^T, \delta\mathbf{x}_r^T]^T$ is given by:

$$\begin{bmatrix} \delta\mathbf{t}_k \\ \delta\mathbf{x}_r \end{bmatrix} \sim \mathcal{N}\left(\begin{bmatrix} \delta\mathbf{t}_k \\ \delta\mathbf{x}_r \end{bmatrix}, \begin{bmatrix} \Sigma_{tt} & \Sigma_{tr} \\ \Sigma_{rt} & \Sigma_{rr} \end{bmatrix}\right)$$

*Proof 2.* By applying the *Woodbury-formula*, with *Proposition 1.*, we have the inverse of Covariance:

$$\Sigma = \begin{bmatrix} \Sigma_{tt} & \Sigma_{tr} \\ \Sigma_{rt} & \Sigma_{rr} \end{bmatrix} = \begin{bmatrix} \mathbf{H}_{tt} & \mathbf{H}_{tr} \\ \mathbf{H}_{rt} & \mathbf{H}_{rr} \end{bmatrix}^{-1} =$$

$$\begin{bmatrix} \mathbf{1} & \mathbf{H}_{tr}\mathbf{H}_{rr}^{-1} \\ \mathbf{0} & \mathbf{1} \end{bmatrix} \begin{bmatrix} \mathbf{H}_{tt} + \mathbf{H}_{tr}\mathbf{H}_{rr}^{-1}\mathbf{H}_{rt} & \mathbf{0} \\ \mathbf{0} & \mathbf{H}_{rr} \end{bmatrix} \begin{bmatrix} \mathbf{1} & \mathbf{0} \\ \mathbf{H}_{rr}^{-1}\mathbf{H}_{rt} & \mathbf{1} \end{bmatrix}$$

. Therefore we have $\Sigma_{tt} = \mathbf{H}_{tt} - \mathbf{H}_{tr}\mathbf{H}_{rr}^{-1}\mathbf{H}_{rt} = \bar{\mathbf{H}}_{tt}^{-1}$, which has exactly the same form with the Covariance derived above.

*2) Experimental Results:* We provide some experimental results to further show the uncertainty estimation, shown in Fig. 1. Generally, the localization error is coherent with the magnitude of covariance corresponding to the translational part. Another interesting observation is that, according to the excerpts, when total number of matched features or matched prior landmarks is insufficient, the uncertainty of estimation tends to be large.

### D. Other Implementation Details

*1) Inertial initialization:* To properly incorporate the inertial information, the system should initialize the bias terms $\mathbf{b}_a, \mathbf{b}_g$ and the rotation between map frame and inertial frame $\mathbf{R}_{WG}$. Here we generally follow the scheme proposed in [3] to initialize parameters relevant to inertial measurements. The difference is that we focus on a localization problem, thus the absolute scale is observable and the $\mathbf{R}_{WG}$ can also be estimated in the mapping procedure. Therefore the only parameters to optimize are $\mathbf{b}_a, \mathbf{b}_g$. When the estimated $\mathbf{b}_a$ between two optimization is close, we consider the initialization converges and the inertial constraints are then added to the localization system.

*2) Handling Localization Failure:* If inertial measurement is available, our system continues using the state prediction with IMU for about 0.5s without re-localization. This is because the re-localization is more time consuming and initializing the inertial parameters would take a few frames. On the contrary, for the vision-only solution, we directly enter the re-localization stage.

### E. Detailed Timing Results

We report the detailed timing results in Tab. I, where the timing results of Pred and Opt correspond to the results in the manuscript. In general, Ours(V+I+L) achieves a real time performance in single thread (28.8 ms on V201 and 38.0ms on V103), while Ours(V+P) can be a little bit time-consuming (37.4 ms on V201 and 54.9 ms on V103).

TABLE I: Runtime comparison on V201 (top) and V103 (bottom) for different modules: Re-localization (Re-Loc), global feature extraction (Feat(G)), local feature extraction (Feat(L)), State Prediction (Pred), instant localization (Track), temporal landmark verification (Update), temporal landmark generation (Creation) and windowed optimization (Opt). Modules marked with (*) are only called when necessary. The proposed association step and optimization strategy do improve the efficiency of the complete system.

| Modules | Ours(V+P) | Ours(V) | Ours(V+I) | Ours(V+I+L) |
|---|---|---|---|---|
| Feat(G)* | 29.1 ± 0.0 | | | |
| Re-Loc* | 105.0 ± 0.0 | | | |
| Feat(L) | 17.7 ± 0.3 | | | |
| Pred | 0.2 ± 0.0 | 0.2 ± 0.0 | 0.2 ± 0.0 | 0.2 ± 0.0 |
| Track | 9.1 ± 0.2 | 9.1 ± 0.2 | 8.7 ± 0.1 | 8.6 ± 0.1 |
| Update | 0.4 ± 0.0 | 0.4 ± 0.0 | 0.4 ± 0.0 | 0.4 ± 0.0 |
| Creation* | 9.3 ± 9.6 | 2.5 ± 0.9 | 2.4 ± 0.7 | 2.4 ± 0.7 |
| Opt* | 50.4 ± 22.1 | 49.4 ± 24.2 | 23.4 ± 6.9 | 8.0 ± 1.9 |
| Total | 37.4 ± 0.3 | 36.2 ± 0.2 | 31.4 ± 0.2 | 28.8 ± 0.1 |
| **Modules** | **Ours(V+P)** | **Ours(V)** | **Ours(V+I)** | **Ours(V+I+L)** |
| Feat(G)* | 29.9 ± 0.0 | | | |
| Re-Loc* | 141.6 ± 0.0 | | | |
| Feat(L) | 16.9 ± 0.6 | | | |
| Pred | 0.1 ± 0.0 | 0.1 ± 0.0 | 0.1 ± 0.0 | 0.1 ± 0.0 |
| Track | 15.7 ± 4.1 | 15.9 ± 0.2 | 15.2 ± 0.1 | 15.0 ± 0.1 |
| Update | 0.5 ± 0.0 | 0.4 ± 0.0 | 0.4 ± 0.0 | 0.4 ± 0.0 |
| Creation* | 4.9 ± 1.4 | 1.2 ± 0.3 | 1.2 ± 0.3 | 1.2 ± 0.3 |
| Opt* | 38.6 ± 11.3 | 33.3 ± 9.1 | 13.1 ± 4.0 | 4.8 ± 1.0 |
| Total | 54.9 ± 5.1 | 51.1 ± 0.3 | 41.6 ± 0.3 | 38.0 ± 0.3 |



\end{document}